\documentclass[runningheads]{llncs}
\usepackage{color}
\usepackage{graphicx}
\usepackage{booktabs}
\usepackage{array}
\usepackage{makecell}
\usepackage[justification=centering]{caption}
\usepackage{subcaption}
\usepackage{indentfirst}
\usepackage[misc]{ifsym}
\usepackage{multicol}
\usepackage{longtable}
\usepackage{hyperref}
\makeatletter
\newcommand\tabcaption{\def\@captype{table}\caption}
\newcommand\figcaption{\def\@captype{figure}\caption}

\def\@fnsymbol#1{\ensuremath{\ifcase#1\or +\or * \fi}}
\makeatother

\begin{document}

\title{Visual Realism Assessment for Face-swap Videos}
%
%
\author{Xianyun Sun\inst{1}\thanks{This work is done while Xianyun Sun is an intern at CASIA.}, Beibei Dong\inst{2}, Caiyong Wang\inst{1}, Bo Peng\inst{2}\thanks{Bo Peng is the corrseponding author.}, Jing Dong\inst{2}}
%
\authorrunning{X. Sun et al.}
%
\institute{Beijing Key Laboratory of Robot Bionics and Function Research, Beijing University of Civil Engineering and Architecture, Beijing, P.R. China 
\email{sunxianyun@stu.bucea.edu.cn},
\email{wangcaiyong@bucea.edu.cn} \\ \and
Center for Research on Intelligent Perception and Computing, Institute of Automation, Chinese Academy of Sciences, Beijing, P.R. China
\email{dongbeibei2022@ia.ac.cn},
\email{bo.peng@nlpr.ia.ac.cn}, 
\email{jdong@nlpr.ia.ac.cn}
%
}
\maketitle              
\begin{abstract}
Deep-learning-based face-swap videos, also known as deepfakes, are becoming more and more realistic and deceiving. The malicious usage of these face-swap videos has caused wide concerns. The research community has been focusing on the automatic detection of these fake videos, but the assessment of their visual realism, as perceived by human eyes, is still an unexplored dimension. Visual realism assessment, or VRA, is essential for assessing the potential impact that may be brought by a specific face-swap video, and it is also important as a quality assessment metric to compare different face-swap methods. In this paper, we make a small step towards this new VRA direction by building a benchmark for evaluating the effectiveness of different automatic VRA models, which range from using traditional handcrafted features to different kinds of deep-learning features. The evaluations are based on a recent competition dataset named DFGC-2022, which contains 1400 diverse face-swap videos that are annotated with Mean Opinion Scores (MOS) on visual realism. Comprehensive experiment results using 11 models and 3 protocols are shown and discussed. We demonstrate the feasibility of devising effective VRA models for assessing face-swap videos and methods. The particular usefulness of existing deepfake detection features for VRA is also noted. The code can be found at \url{https://github.com/XianyunSun/VRA.git}.

\end{abstract}
\section{Introduction}
\label{sec:intro}

\begin{figure}[t]
\centering
	\begin{minipage}{0.4\linewidth}
		\centering
		\includegraphics[width=\linewidth]{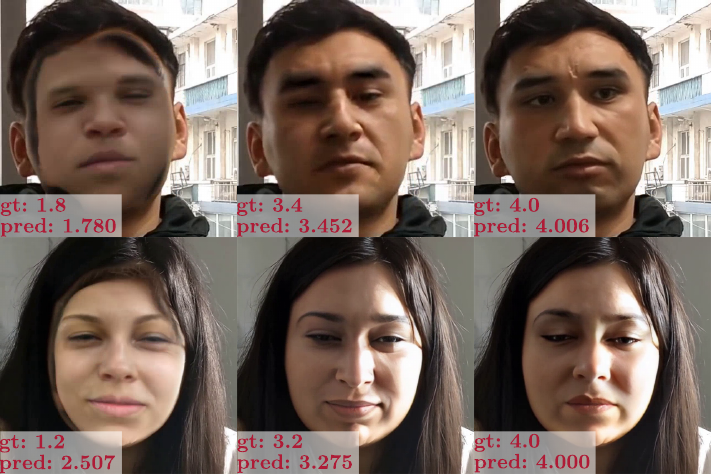}
        \subcaption{}
        \label{fake1}
	\end{minipage} \hfill
 \begin{minipage}{0.28\linewidth}
		\centering
		\includegraphics[width=\linewidth]{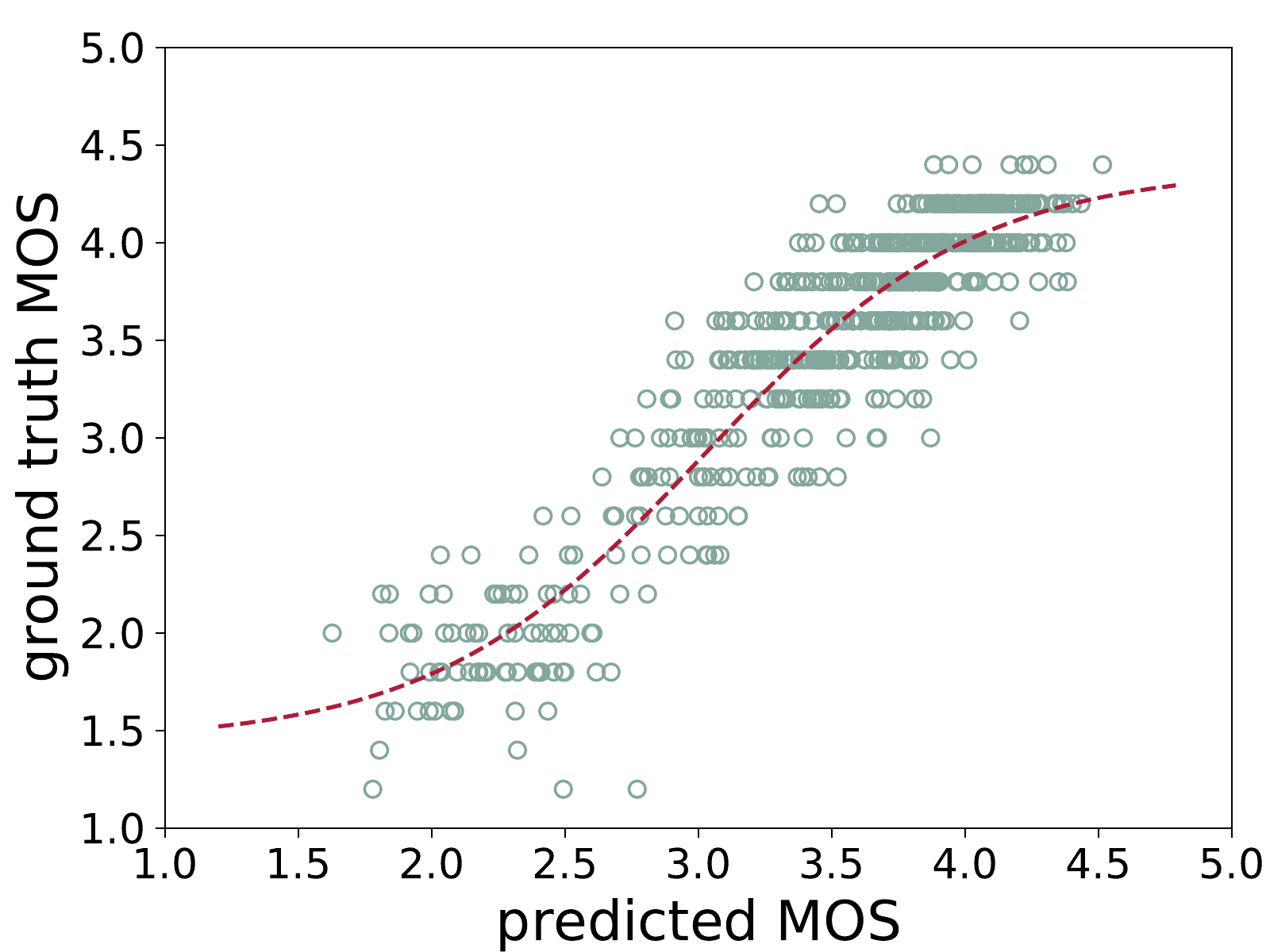}
        \subcaption{}
        \label{predict-subid}
	\end{minipage}\hfill
 \begin{minipage}{0.28\linewidth}
	\centering
	\includegraphics[width=\linewidth]{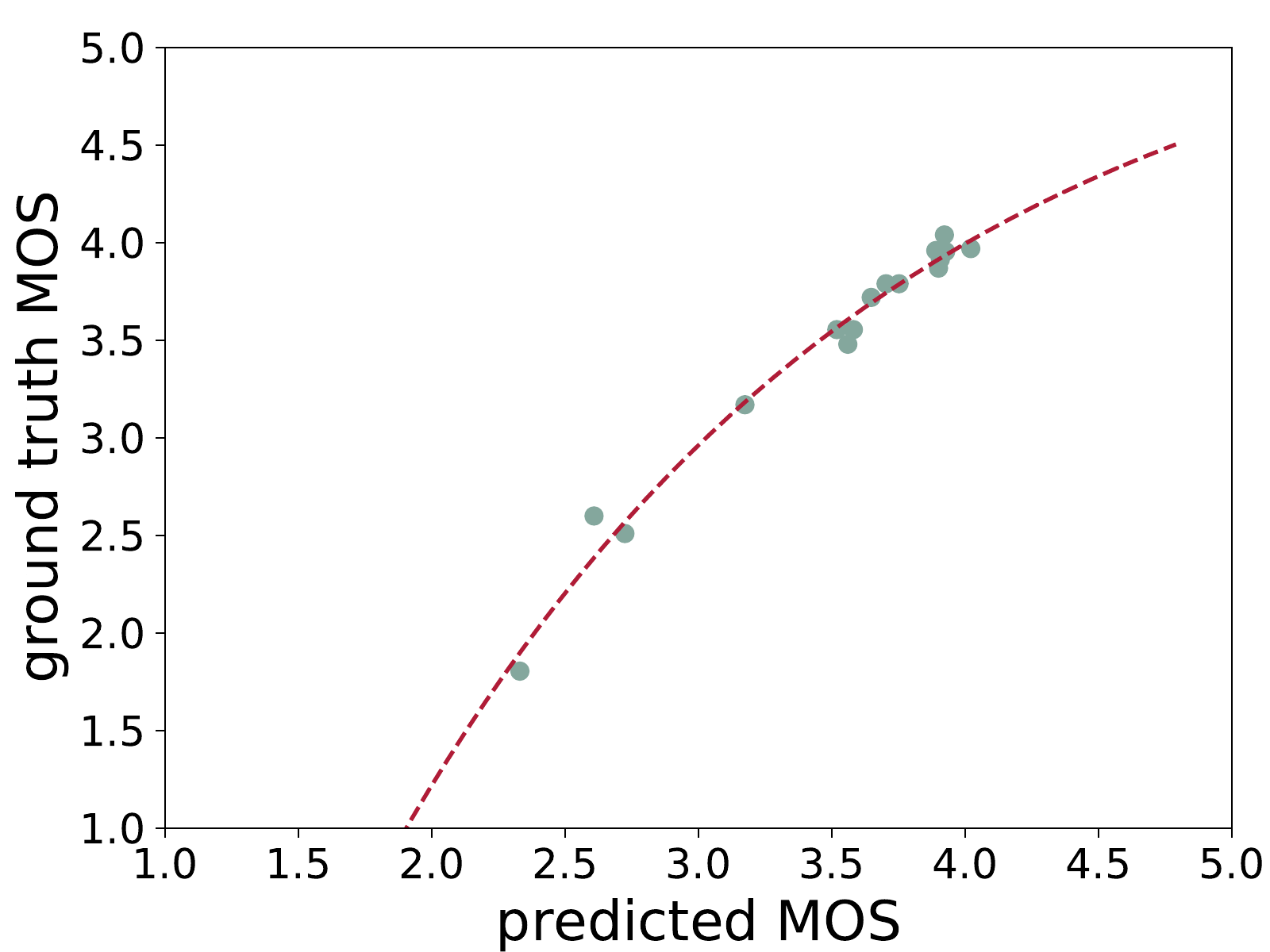}
    \subcaption{}
    \label{predict-method}
\end{minipage}
\caption{(a) Face-swap videos with different degrees of realism, annotated with the ground truth MOS (\emph{gt}) vs predicted MOS (\emph{pred}) by the DFGC-1st VRA model. (b) and (c) are scatter plots with the fitted logistic curves (see Subsection \ref{protocols}). (b) is the video-level plot and (c) is the method-level plot.} 
\label{demo}
\end{figure}

Face-swap videos, as the name indicates, are videos in which the appearance of a face is manipulated using computer programs (especially deep learning based methods) so that audiences may recognize the face as another individual. This technology has contributed a lot to filming and other entertainment industries, yet holding a high risk of being abused. The detection methods against face-swap videos, or deepfakes, have improved a lot with intense attention being drawn \cite{juefei2022countering}.
Since the ultimate goal of face-swapping is to serve human viewers, subjective realism assessment could play a critical role not only in estimating the influence of fake videos on social networks, but also in evaluating the performance of face-swapping models during their development. \par

Several studies have been carried out to explore the subjective opinions on the persuasiveness of deepfake media. Deep models such as MOSNet \cite{mos-net} and MOSA-Net \cite{mosa-net} are developed for assessing the naturalness of converted speeches. Compared with deepfake audios, relatively fewer studies have been carried out on deepfake images or videos.

Nightingale et al. \cite{ai}, \cite{synface} conduct subjective evaluations on StyleGAN2-generated images and find that the synthetic faces are indistinguishable from and even more trustworthy than real faces. Korshunov and Marcel \cite{human-vs-machines} conduct a subjective study on face-swap videos from the DFDC dataset \cite{dfdc} and find that human perception is very different from the machine perception, and they are both successfully but in different ways fooled by deepfakes. All these studies, however, only demonstrate human performance in deepfake detection, with none of them providing any quantitative method to estimate the realism degree of deepfakes. A model proposed in \cite{tian2022generalized} is trained to predict subjective quality for GAN-generated facial images, which is the only model of its kind to the best of our knowledge. There is an obvious vacant position for models assessing the visual realism of deepfake videos. \par

Here in this paper, we build the first visual realism assessment (VRA) benchmark for face-swap videos as an attempt to fill this gap. In our proposed method, models from related fields are employed as feature extractors, with support vector regression (SVR) as the regressor mapping features to a predicted subjective realism score. Fig. \ref{fake1} shows a demo of some frames from fake videos with the ground-truth mean opinion score (MOS) and predicted MOS. Fig. \ref{predict-subid} and Fig. \ref{predict-method} show a general view of the correlation between the prediction and the groundtruth on the DFGC-2022 dataset, in video-level and method-level  (i.e. face-swap methods assessed) respectively.\par

In the following parts, Section \ref{dataset intro} gives a brief overview of the DFGC-2022 dataset, on which our work is based. Section \ref{method} introduces the proposed VRA method. Experiment details and results are discussed in Section \ref{experiment}, and Section \ref{conclusion} summarizes this work.

\section{Dataset Analysis} \label{dataset intro}

\begin{figure}
\centering
\begin{minipage}{0.45\linewidth}
\small
    \begin{tabular}{rccc}\hline
        \textbf{subset} & \textbf{\makecell[c]{annotated \\fake clips}} & \textbf{\makecell[c]{facial-\\ids}} & \textbf{\makecell[c]{submit-\\ids}}\\ \hline
        \textbf{C1} & 240  & 20 pairs & 6  \\
        \textbf{C2} & 520  & 20 pairs & 13 \\
        \textbf{C3} & 640  & 20 pairs & 16 \\ \hline
    \end{tabular}
    \tabcaption{ Details of C1, C2, and C3 subset in DFGC-2022}
    \label{dataset} 
\end{minipage} \hfill
\begin{minipage}{0.54\linewidth}
	\includegraphics[width=\linewidth]{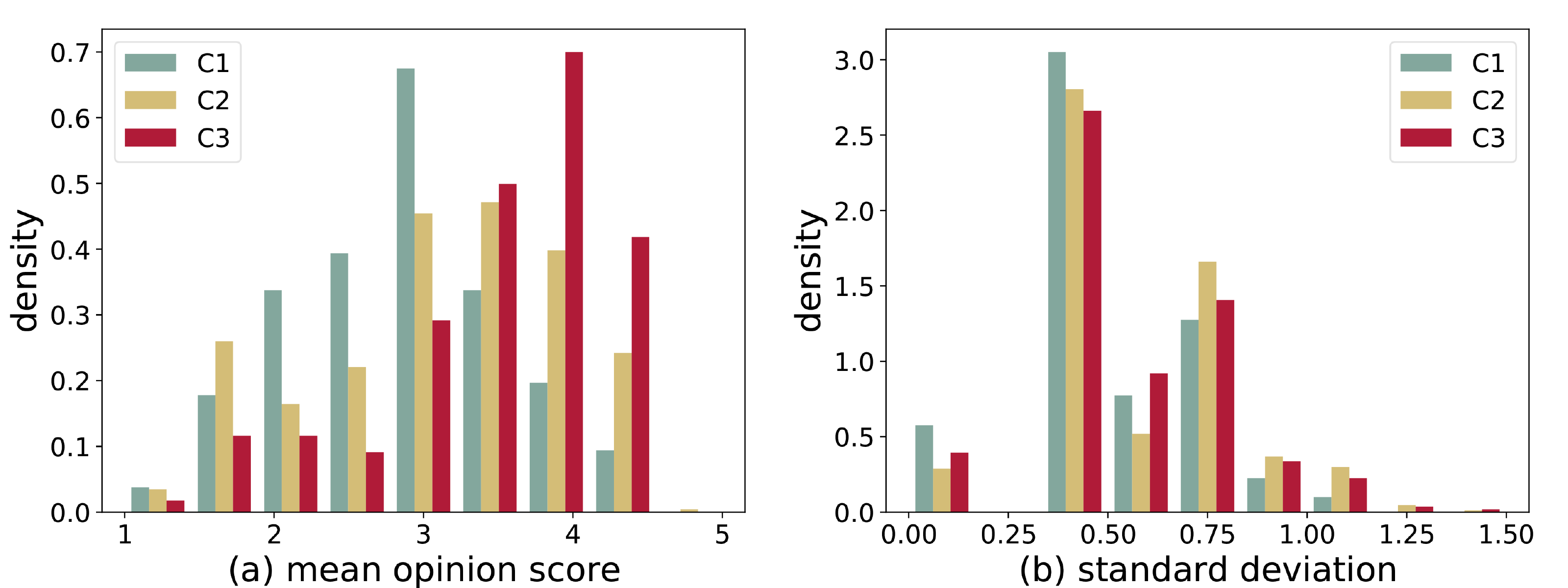}
    \figcaption{Histograms of the mean and standard deviation of each video's realism rating in DFGC-2022.} 
    \label{mean_mos}
\end{minipage}
\vspace{-30pt}
\end{figure}

Originated from the Second DeepFake Game Competition (DFGC) \cite{dfgc} held with the IJCB-2022 conference, the DFGC-2022 dataset contains a total of 2799 face-swap videos and 1595 real videos, all about 5s in length. Fake clips in the dataset are generated by various face-swap methods (e.g. DeepFaceLab \cite{deepfacelab}, SimSwap \cite{simswap}, FaceShifter \cite{faceshifter}) and post-processing operations, and they are submitted by the participants through three separate submission sessions, i.e., C1, C2, and C3. This forms three subsets, with their details shown in Table \ref{dataset}.  Each submission is associated with a submit-id and contains 80 swap videos for 20 pairs of facial IDs. The fake clips from the same submit-id are deemed to be created by the same method or process. 40 clips in each submission are annotated by 5 human raters independently in the aspect of video realism, apart from some other aspects. The rating is from 1 (very bad) to 5 (very good). The mean opinion score (MOS) and the standard deviation of each video's ratings are calculated and their distributions are shown in Fig.\ref{mean_mos}.

\section{VRA Methods} \label{method}
In this section, we will go through the workflow of the VRA methods. First, we crop the face area from each frame in the data pre-processing step. Then, per-frame features are extracted using existing handcrafted or deep-learning models. And finally, the per-frame features are fused into video-level feature, which goes through a feature selection step before it is used to regress the video realism score. 
Here we follow the classical video quality assessment workflow \cite{videval} to construct our own, which is in contrast to learning end-to-end deep models for VRA, e.g. using LSTM \cite{lstm} and GRU \cite{gru} models. This is because deep models heavily rely on the amount of  training data, which is not suitable in our case, considering that there are only several hundreds of annotated training videos in the DFGC-2022 dataset. \par


\subsection{Data Pre-processing}
The videos in DFGC-2022 dataset have the resolution of $1080 \times 1920$, with the speaker's face taking over less than 25\% of the area. Since VRA for face-swap  should focus more on the facial area than the backgrounds, we crop each video according to detected face bounding boxes. 
For the fairness of comparing  different face-swap methods, the face detection is only performed on the original target videos, and the result boxes are shared by all face-swap videos that originate from the same target video. 

Specifically, we first enlarge the detected boxes by 1.3 times to include the full head region. We then obtain the smallest box that encapsulate all face boxes in the video and use it to crop all the video frames. This cropping strategy prevents the jittering of consecutive cropped frames.    
Cropped videos shrunk to about $600 \times 600$, which is also beneficial for the time efficiency of following processes. 

\subsection{Feature Extraction}
For feature extractors, we employ several representative models from the subjective image/video quality assessment (I/VQA), image recognition, face recognition, and deepfake detection fields, with the consideration of potential feature sharing between VRA and these tasks. Table \ref{sum} summarises the included models. 
The \emph{original} part in the \emph{feats dim} column refers to the dimension of the original video-level features extracted by each model, and the \emph{selected} part denotes the dimension after our feature selection step. 

\begin{table}[htb]
\setlength{\abovecaptionskip}{0cm}  
\setlength{\belowcaptionskip}{-0.2cm} 
    \centering
    \tabcolsep=5pt
    \caption{\centering Overview of the feature extraction models.}
    \begin{tabular}{r|cccc}
    \hline
        \textbf{Model} & \textbf{Original Task} & \multicolumn{2}{c}{\textbf{feats dim}} & \textbf{Training Data} \\
        \cmidrule(lr){3-4}
        ~ & ~  & \textbf{original} & \textbf{selected} & ~ \\ 
        \hline
        \textbf{BRISQUE} & IQA & 72 & 72 & handcrafted  \\
        \textbf{GM-LOG} & IQA & 80 & 80 & handcrafted \\
        \textbf{FRIQUEE} & IQA &  1120 & 1120 & handcrafted \\
        \textbf{TLVQM} & VQA &  75 & 75 & handcrafted \\
        \textbf{V-BLIINDS} & VQA &  46 & 46 & handcrafted \\
        \textbf{VIDEVAL} & VQA &  60 & 60 & handcrafted \\
        \textbf{ensemble} & VQA &  3229 & 240 & handcrafted \\
        \textbf{ResNet50} & image recognition &  4096 & 160 &  ImageNet \\
        \textbf{VGG-Face} & face recognition &  8192 & 280 & VGG-Face \\
        \textbf{DFDC-ispl} & deepfake detection & 3584 & 100 & FF++, DFDC \\
        \textbf{DFGC-1st} & deepfake detection &  11264 & 260 & 9 deepfake datasets \\
        \hline
    \end{tabular}
    \label{sum}
\end{table}

\textbf{IQA Features.} 
BRISQUE \cite{brisque} is a typical IQA model under the natural scene statistics (NSS) framework, adopting mean subtracted contrast normalized (MSCN) coefficients as its band-pass filter. 
The FRIQUEE model \cite{friquee} further extends the application of NSS model from gray scale to multiple color spaces including RGB, LAB and LMS.
GM-LOG \cite{gmlog} uses isotropic differential operators in replacement of band-pass transforms, including the Gradient Magnitude (GM) and Laplacian of Gaussian (LOG) operators. 

\textbf{VQA Features.}
Different from IQA features that only focus on single frames, VQA features also represent the temporal information. 
TL-VQM \cite{tlvqm} 
includes statistical features of the motion vectors between every two consecutive frames. 
V-BLIINDIS \cite{vbliinds} also includes features of the motion vectors and includes DCT features extracted from frame differences. 

VIDEVAL \cite{videval} is a SOTA VQA model which packs up features from BRISQUE, GMLOG, FRIQUEE and TLVQM and employs an additional feature selection process. 
Inspired by VIDEVAL, we propose a new \emph{ensemble} model that extends VIDEVAL's feature candidates to also include features from V-BLIINDS and the handcraft features in RAPIQUE \cite{RAPIQUE}. 
Similarly, the feature selection process is conducted, as will be introduced in Section \ref{regression}.\par


\textbf{General Image Recognition Features.}
In existing VQA literature, it has been shown that features extracted by general-purpose image recognition models like VGG \cite{vgg} and ResNet \cite{resnet} pre-trained on ImageNet can be potential video quality indicators with an additional regressor on top. This makes it a natural choice for us to also include the image recognition features in our evaluation. A pre-trained ResNet-50 is adopted, and we resize the images according to its input requirements.\par

\textbf{Face Recognition Features.}
VGG-Face \cite{vggface} are selected as the representative face recognition model for feature extraction. VGG-Face achieves a remarkable face recognition accuracy using a VGG-19 model finetuned on the VGG-Face dataset including 2622 identities. We select it for its simplicity and high performance.

\textbf{Deepfake Detection Features.}
Since our VRA benchmark is based on the DFGC-2022 dataset, the 1st-place solution \cite{url_DFGC1st} in DFGC-2022 detection track (referred to as DFGC-1st) is a natural candidate for evaluation. Two ConvNext at different epochs and a Swin-Transformer are employed in this solution, and they are trained on an abundant collection of 9 deepfake  datasets with data augmentation and two-class classification loss. Note that the DFGC-2022 dataset itself is not in the training data of this model.

As a comparison, we also include a top 2\% solution from the ISPL team in the DFDC challenge \cite{ispl} (referred to as DFDC-ispl). This solution employs a single EfficientNet with extra attention blocks, which is trained on two datasets, i.e., FaceForensics++ \cite{ff++} and DFDC. 

\subsection{Realism Score Regression}\label{regression}

\textbf{Video-level Feature Fusion.}
Apart from the VQA features, i.e., TLVQM, V-BLIINDS, VIDEVAL and ensemble, which are already extracted as video-level features following their original fusion designs, the rest are per-frame features and need to be fused to video-level features. 
With frame features $f_1, f_2, ..., f_n$ extracted from $n$ sampled frames, average pooling ($f_{mean}$) and standard deviation pooling ($f_{std}$) are the two most popular feature aggregation methods in the VQA field, which are also adopted in our work. 
Note that $f_{mean}$ and $f_{std}$ each has the same feature dimension as the frame features, and they are concatenated to form the video-level features. Take the ResNet50 model in Table \ref{sum} as example, the dimension of frame features extracted by the model is 2048, then the fused video-level feature dimension becomes 4096 after concatenating the mean and std.

\begin{figure}[htb]
\centering
\includegraphics[width=0.7\textwidth]{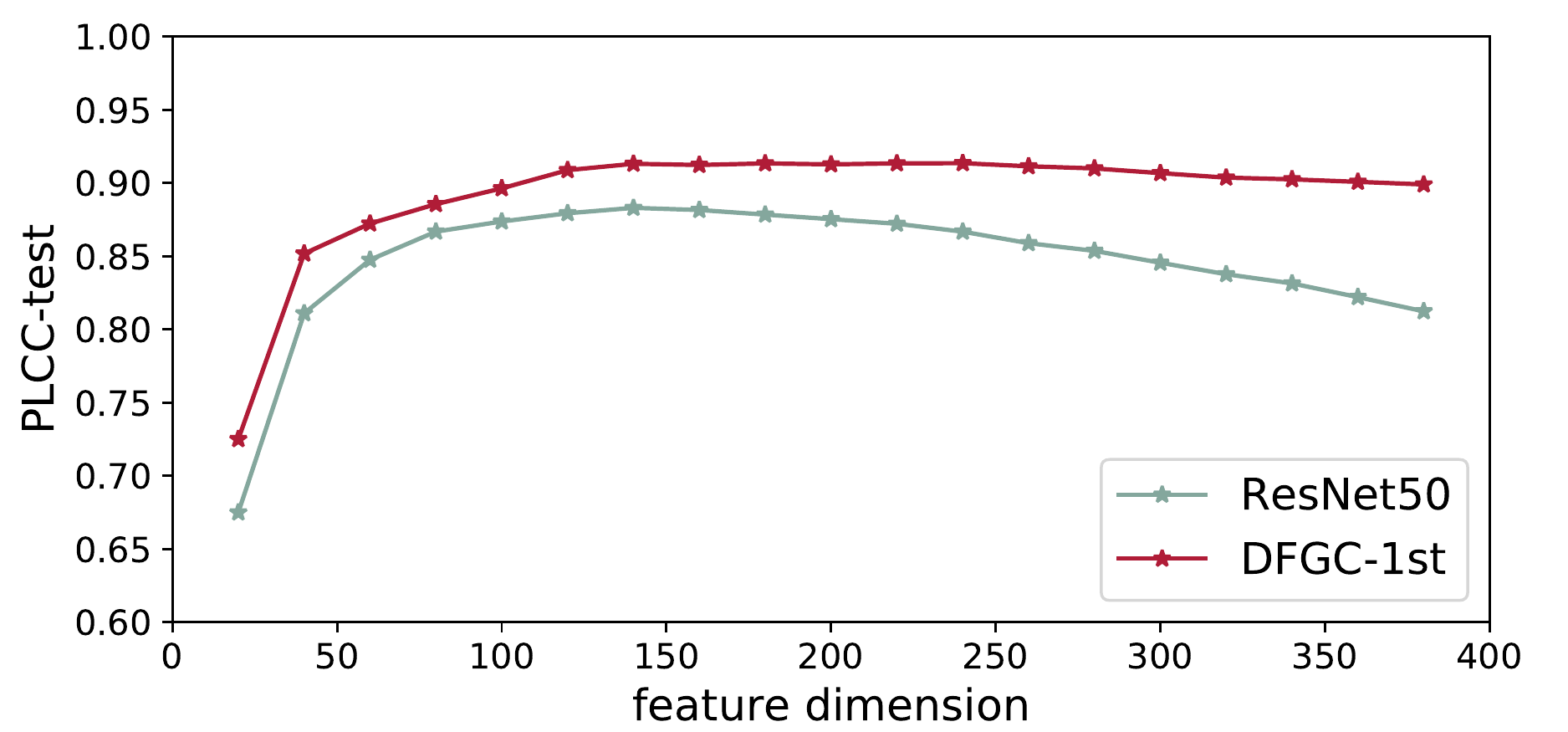}
\caption{PLCC under different selected feature dimensions.} \label{select num}
\end{figure}

\textbf{Feature Selection.}
The performance and efficiency of our regressor, which is a SVR here, drop prominently when feature dimension grows too large, indicating a need for feature selection. Fig. \ref{select num} shows how the accuracy of ResNet50 and DFGC-1st changes with the dimension of selected features. Our feature selections are conducted for the ensemble, ResNet50, VGG-Face, DFDC-ispl, and DFGC-1st models, as their original feature dimensions are relatively high, as shown in Table \ref{sum}. \par

Following the VQA work presented in \cite{videval}, we implement a similar two-stage feature selection strategy. In both selection stages, feature importance is ranked by a SVR with the linear kernel. In the first stage, the optimal number of features $k$ is selected by gird-search over the range of total feature dimensions in a step of 20. The $k$ giving the best average PLCC  in  10 random train-test iterations is chosen. Each feature extraction model has its own optimal $k$, as shown in the \textit{selected} sub-column of Fig. \ref{sum}. In the second stage, 100 iterations are preformed with the optimal $k$, resulting in 100 subsets of chosen features. The frequency of each feature being selected over these iterations is recorded, and the top-$k$ most frequent features are selected as the final selected features. More details can be found in \cite{videval}. \par

\textbf{Score Regression.}
With each model's selected features as the input, support vector regression (SVR) models are trained to regress the groundtruth MOS of video realism, using L2 loss. For this score regression step, we use the SVR model with RBF kernel, and set its hyper-parameters  $C$ and $\gamma$ by grid-search using a random 20\% of the  training data as the validation set. Finally, the regressor is trained again on the whole training set with the searched hyper-parameters.

\begin{table}[htbp]
\caption{Performance comparison of VRA models}
\label{general result}
\begin{subtable}{\linewidth}
    \centering
    \tabcolsep=10pt
    \caption{ Performance under video level facial-id split}
    \begin{tabular}{r|ccc}
    \textbf{Metric} & \textbf{SRCC↑(std)} &  \textbf{PLCC↑(std)}  & \textbf{RMSE↓(std)} \\ \hline 
        \textbf{BRISQUE} & 0.2646(.104) & 0.4185(.124) & 0.6473(.055)  \\ 
        \textbf{GM-LOG} & 0.4324(.097) & 0.5630(.088) & 0.5907(.053)  \\ 
        \textbf{FRIQUEE} & 0.5281(.084) & 0.6926(.078) & 0.5134(.059) \\ 
        \textbf{TLVQM} & 0.3988(.081) & 0.5586(.096) & 0.5923(.058) \\ 
        \textbf{V-BLIINDS} & 0.4042(.114) & 0.6251(.123) & 0.5502(.071) \\ 
        \textbf{VIDEVAL} & 0.3277(.124) & 0.4521(.104) & 0.6376(.054) \\ 
        \textbf{ensemble} & \underline{0.6364(.063)}  & \underline{0.7979(.052)} & \underline{0.4298(.052)} \\
        \textbf{ResNet50} & 0.6006(.083) & 0.7827(.059) & 0.4420(.049) \\
        \textbf{VGGFace} & 0.5814(.111) & 0.7710(.078) & 0.4486(.054) \\ 
        \textbf{DFDC-ispl} & 0.5641(.092) & 0.7868(.061) & 0.4380(.047) \\ 
        \textbf{DFGC-1st} & \textbf{0.7952(.051)} & \textbf{0.8975(.028)} &  \textbf{0.3132(.030)} \\ \bottomrule
    \end{tabular}
    
\end{subtable}

\begin{subtable}{\linewidth}
\centering
\tabcolsep=10pt
\caption{ Performance under video level submit-id split}
    \begin{tabular}{r|ccc}
    \textbf{Metric} & \textbf{SRCC↑(std)} & \textbf{PLCC↑(std)} & \textbf{RMSE↓(std)} \\ \hline        
    \textbf{BRISQUE} & 0.5379(.202) & 0.5803(.198) & 0.4208(.135)  \\ 
    \textbf{GM-LOG} & 0.5160(.229) & 0.5657(.226) & 0.4152(.114) \\ 
    \textbf{FRIQUEE} & 0.6481(.165) & 0.6928(.175) & 0.3536(.082) \\ 
    \textbf{TLVQM} & 0.5593(.195) & 0.6165(.203) & 0.3097(.096) \\ 
    \textbf{V-BLIINDS} & 0.4851(.235) & 0.5316(.247) & 0.4166(.096) \\ 
    \textbf{VIDEVAL} & 0.5438(.201) & 0.6014(.202) & 0.4047(.119) \\ 
    \textbf{ensemble} & 0.7211(.142) & 0.7628(.152) & 0.3020(.048) \\
    \textbf{ResNet50} & 0.7423(.126) & 0.7868(.132)& 0.2905(.043) \\ 
    \textbf{VGGFace} & \underline{0.7673(.100)}& 0.7922(.113) & 0.3049(.094) \\ 
    \textbf{DFDC-ispl} & 0.7582(.115) & \underline{0.8009(.129)} & \underline{0.2825(.050)} \\ 
    \textbf{DFGC-1st} & \textbf{0.8081(.096)} & \textbf{0.8356(.106)} & \textbf{0.2540(.037)}\\ 
    \bottomrule
    \end{tabular}
\end{subtable}

\begin{subtable}{\linewidth}
\centering
\tabcolsep=10pt
\caption{ Performance under method level submit-id split}
    \begin{tabular}{r|ccc}
    \textbf{Metric} & \textbf{SRCC↑(std)} & \textbf{PLCC↑(std)} & \textbf{RMSE↓(std)} \\ \hline
    \textbf{BRISQUE} & 0.6906(.453) & 0.7687(.453) & 0.2730(.226) \\ 
    \textbf{GM-LOG} & 0.6970(.476) & 0.7500(.472) & 0.2887(.209) \\ 
    \textbf{FRIQUEE} & 0.8120(.347) & 0.8712(.312) & 0.2073(.181) \\ 
    \textbf{TLVQM} & 0.7170(.428) & 0.7749(.432) & 0.2578(.175) \\ 
    \textbf{V-BLIINDS} & 0.6833(.510) & 0.7263(.506) & 0.2428(.146) \\ 
    \textbf{VIDEVAL} & 0.7633(.418) & 0.8109(.383) & 0.2696(.210) \\ 
    \textbf{ensemble} & 0.8756(.271) & 0.9168(.276) & \underline{0.1281(.107)} \\
    \textbf{ResNet50} & 0.8370(.342) & 0.9048(.295) & 0.1362(.106) \\ 
    \textbf{VGGFace} & \underline{0.9496(.144)} & \textbf{0.9746(.056)} & 0.1840(.189) \\
    \textbf{DFDC-ispl} & 0.8656(.275) & 0.9407(.250) & 0.1401(.097)\\ 
    \textbf{DFGC-1st} & \textbf{0.9556(.129)} & \underline{0.9715(.082)} & \textbf{0.1141(.105)}  \\
    \bottomrule
    \end{tabular}
\end{subtable}
\end{table}

\section{Experiment Results} \label{experiment}
\subsection{Evaluation Protocols and Metrics} \label{protocols}
Since C3 is the session with the largest number of submissions, as shown in Table \ref{dataset}, we train our models exclusively on the C3 subset of DFGC-2022. We report both intra-subset and inter-subset performances, where the former trains on a portion of C3 videos and tests on the rest C3 videos, while the latter trains on C3 and tests on C1 and C2. 

For the intra-subset evaluation, we report model performances using three different protocols: the \emph{video level facial-id} split, the \emph{video level submit-id} split, and the \emph{method level submit-id} split. In the facial-id split, 4 out of 20 ID pairs (128 out of 640 videos) are chosen as the test set, and the rest are the train set. In the submit-id spits, 3 out of 16 submit-IDs (120 out of 640 videos) are chosen as the test set. To reduce the impact of randomness, 100 train-test iterations are preformed with different choices of facial-ids or submit-ids across iterations. A random seed equals to the iteration number is set to ensure the uniformity of splits when testing different models.  \par

Different from the video level protocols that calculate prediction accuracy for videos, the method level protocol aims to evaluate the overall quality of different face-swap methods with respect to the realism of their created videos. For method level evaluation, the groundtruth method MOS is calculated by the average of groundtruth MOS of videos in the same submit-id, and the predicted method MOS is the average of predicted MOS of these videos. \par

For the inter-subset evaluations, the models are trained on all C3 videos and tested on all C1 or C2 videos. It is a more challenging protocol that can reflect the generalization ability of evaluated models. This is because C2 and C1 videos are created by different face-swap methods from C3 and their MOS has different distributions. Note that, in the inter-subset setting, the selected features and hyper-parameters are all the same from those in the intra-subset setting, meaning that the models are not fine-tuned from sets to sets.

Following the VQA literature, SRCC (Spearman rank-order correlation coefficient), PLCC (Pearson linear correlation coefficient) and  RMSE (root mean square error) are employed as evaluation metrics in our benchmark. The average value over all testing iterations is reported to reflect model performances, and the standard deviation is also shown, which can imply the robustness of the models. As suggested in \cite{seshadrinathan2010study} and \cite{videval}, a nonlinear logistic function with four parameters is fitted to the predicted MOS before calculating the final metrics to improve prediction accuracy. 

\begin{table}[htbp]
\setlength{\abovecaptionskip}{0cm}
    \tabcolsep=5pt
    \caption{Inter-subsets evaluation results}
    \label{inter subset}
    \begin{subtable}{\linewidth}
    \centering
    \caption{Training on C3 and testing on C1}
    \begin{tabular}{r|cccccc}
    \hline
        ~ & \multicolumn{2}{c}{\textbf{SRCC↑}} & \multicolumn{2}{c}{\textbf{PLCC↑}} & \multicolumn{2}{c}{\textbf{RMSE↓}} \\ 
        \cmidrule(lr){2-3} \cmidrule(lr){4-5} \cmidrule(lr){6-7} 
        \textbf{model} & \textbf{video} & \textbf{method} & \textbf{video} & \textbf{method} & \textbf{video} & \textbf{method} \\  \hline
        \textbf{random} & \multicolumn{2}{c}{-0.0091} & \multicolumn{2}{c}{-0.0084} & \multicolumn{2}{c}{1.3727} \\
        \textbf{ResNet50} & 0.2384 & 0.4857 & 0.2978 & 0.7818 & 0.6834 & 0.3158 \\
        \textbf{VGG Face} & 0.2512 & 0.6571 & 0.2939 & 0.8315 & 0.6843 & 0.2813\\ 
        \textbf{DFDC-ispl} & 0.3367 & \textbf{0.9428} & 0.3595 & \textbf{0.9406} & 0.6680 & 0.1719 \\ 
        \textbf{DFGC-1st} & \textbf{0.3743} & 0.6571 & \textbf{0.4222} & 0.7818 & \textbf{0.6489} & \textbf{0.3158} \\ \bottomrule
    \end{tabular}\label{inter-c1}
    \end{subtable}

    \begin{subtable}{\linewidth}
    \centering
    \caption{Training on C3 and testing on C2}
    \begin{tabular}{r|cccccc}
    \hline
        ~ & \multicolumn{2}{c}{\textbf{SRCC↑}} & \multicolumn{2}{c}{\textbf{PLCC↑}} & \multicolumn{2}{c}{\textbf{RMSE↓}} \\ 
        \cmidrule(lr){2-3} \cmidrule(lr){4-5} \cmidrule(lr){6-7} 
        \textbf{model} & \textbf{video} & \textbf{method} & \textbf{video} & \textbf{method} & \textbf{video} & \textbf{method} \\ \hline
        \textbf{random} & \multicolumn{2}{c}{0.0083} & \multicolumn{2}{c}{0.0083} & \multicolumn{2}{c}{1.411} \\
        \textbf{ResNet50} & 0.4173 & 0.6044 & 0.4027 & 0.7607 & 0.7460 & 0.4582 \\ 
        \textbf{VGG Face} & 0.4522 & 0.6813 & 0.4350 & 0.7580 & 0.7339 & 0.4605 \\ 
        \textbf{DFDC-ispl} & 0.3554 & 0.6978 & 0.3477 & 0.7698 & 0.7642 & 0.4507 \\ 
        \textbf{DFGC-1st} & \textbf{0.5045} & \textbf{0.8846} & \textbf{0.4844} & \textbf{0.9088} & \textbf{0.7130} & \textbf{0.2946} \\ \bottomrule
    \end{tabular}\label{inter-c2}
    \vspace{5pt}
    \end{subtable}
\end{table}

\subsection{Intra-subset Evaluation}
Table \ref{general result} shows a general image comparing different VRA models under our intra-subset evaluation protocols. As can be seen, the DFGC-1st model outperforms the others under nearly all metrics and protocols. The ensemble, VGG-Face, and DFDC-ispl models have the second-rank under some metrics and protocols. While the other handcrafted IQA and VQA models suffer from obliviously lower accuracy. The result of the DFGC-1st model leading the board implies that deepfake detection features may relate most to the VRA problem at hand. \par

Comparing with the video level results, it is clear that the method level counterparts are much more accurate for all models and metrics. This shows that evaluating the realism performance for different face-swap methods are more tractable than that for individual videos. This result is not so unexpected, considering that method level evaluations can average out prediction noises on video instances.

\subsection{Inter-subsets Evaluation}
Table \ref{inter subset} demonstrates the results of the prediction accuracy of models when trained and tested on different data subsets. Handcrafted models are not tested here due to their high computational cost in feature extraction. The DFGC-1st model again surpasses the other models in terms of generalization ability. Although much better than a random guesser (random prediction in $[1, 5]$), all models have a clear performance degradation at video level compared to the intra-subset setting. The situation improves when coming to method level evaluations, but the accuracy gap between intra- and inter-subsets is still obvious. This calls for further study on improving the generalization ability of VRA models.

\begin{table}[htbp]
\setlength{\abovecaptionskip}{0cm}
\setlength{\belowcaptionskip}{-10pt} 
    \centering
    \tabcolsep=5pt
    \caption{Video level performance comparison of popular objective quality metrics on C3}
    \begin{tabular}{r|ccc}
    \hline
        \textbf{metric} & \textbf{SRCC↑} & \textbf{PLCC↑} & \textbf{RMSE↓}\\ \hline
        \textbf{SSIM} &  -0.0814 & 0.1789 & 0.7256  \\
        \textbf{LPIPS} & -0.1312  & 0.2918 & 0.6941 \\
        \textbf{FAST-VQA} & 0.1094 & 0.1104 & 0.9679 \\
        \textbf{\makecell[r]{DFGC-1st \\ detection score}} & 0.2651 & 0.3232 & 0.6867  \\
        \textbf{\makecell[r]{DFGC-1st \\ VRA score}} & 0.8081 & 0.8356 & 0.2540 \\       
        \hline
    \end{tabular}
    \label{subject}
\end{table}

\subsection{Comparison with popular objective quality metrics}
Table \ref{subject} demonstrates the performance of several existing objective quality metrics: SSIM \cite{ssim} and LPIPS \cite{lpips} are commonly used for evaluating deepfake generation models, FAST-VQA \cite{fastvqa} is a SOTA VQA model, the detection score of DFGC-1st reflects the probability of a video being a fake one predicted by the model. FID \cite{fid} cannot be applied here since it is a metric for evaluating the quality of a set of samples instead of a single one. It can be seen that comparing with the VRA scores predicted by DFGC-1st, none of these existing metrics can perform well as a predictor for human perception of deepfake realism. Also, since VRA and anti-detection scores originate from the same DFGC-1st features in this example, the result indicates that our feature selection and regression process play an important role in extracting VRA-related information.

\section{Conclusions}\label{conclusion}
In this paper, we propose a benchmark for the new visual realism assessment (VRA) problem of face-swap videos. This benchmark is based on the DFGC-2022 dataset and includes several models from related fields which are used as feature extractors. An SVR is trained as the regressor to predict realism scores for fake videos. We find that deep features beat most handcrafted ones in this VRA task, with a deepfake detection model trained on diverse datasets, i.e., the DFGC-1st model, achieving the best performance, implying the close relation between deepfake realism assessment and its detection. However, improving VRA's generalization ability under new datasets is still an open problem that requires further research. This work serves as a reference for future studies.

\subsubsection{Acknowledgment}
This work is supported by Beijing Natural Science Foundation under Grant No. 4232037, the National Natural Science Foundation of China (NSFC) under Grants 62272460, U19B2038, 62106015, a grant from Young Elite Scientists Sponsorship Program by CAST (YESS), CAAI-Huawei MindSpore Open Fund, the Pyramid Talent Training Project of Beijing University of Civil Engineering and Architecture (JDYC20220819), the 2023-2025 Young Elite Scientist Sponsorship Program by BAST (BYESS2023130), and the BUCEA Post Graduate Innovation Project (PG2023090).

%
%
%
\nocite{*}
\bibliographystyle{splncs04}
\bibliography{reference}

\begin{thebibliography}{10}
\providecommand{\url}[1]{\texttt{#1}}
\providecommand{\urlprefix}{URL }
\providecommand{\doi}[1]{https://doi.org/#1}

\bibitem{url_DFGC1st}
Dfgc-2022 first-place solution of the detection track.
  \url{https://github.com/chenhanch/DFGC-2022-1st-place}

\bibitem{ispl}
Bonettini, N., Cannas, E.D., Mandelli, S., Bondi, L., Bestagini, P., Tubaro,
  S.: Video face manipulation detection through ensemble of cnns. In: 2020 25th
  International Conference on Pattern Recognition (ICPR). pp. 5012--5019. IEEE
  (2021)

\bibitem{simswap}
Chen, R., Chen, X., Ni, B., Ge, Y.: Simswap: An efficient framework for high
  fidelity face swapping. In: Proceedings of the 28th ACM International
  Conference on Multimedia. pp. 2003--2011 (2020)

\bibitem{gru}
Chung, J., Gulcehre, C., Cho, K., Bengio, Y.: Empirical evaluation of gated
  recurrent neural networks on sequence modeling. arXiv preprint
  arXiv:1412.3555  (2014)

\bibitem{dfdc}
Dolhansky, B., Bitton, J., Pflaum, B., Lu, J., Howes, R., Wang, M., Ferrer,
  C.C.: The deepfake detection challenge (dfdc) dataset. arXiv preprint
  arXiv:2006.07397  (2020)

\bibitem{friquee}
Ghadiyaram, D., Bovik, A.C.: Perceptual quality prediction on authentically
  distorted images using a bag of features approach. Journal of Vision
  \textbf{17}(1),  32--32 (2017)

\bibitem{resnet}
He, K., Zhang, X., Ren, S., Sun, J.: Deep residual learning for image
  recognition. In: Proceedings of the IEEE Conference on Computer Vision and
  Pattern Recognition (CVPR). pp. 770--778 (2016)

\bibitem{fid}
Heusel, M., Ramsauer, H., Unterthiner, T., Nessler, B., Hochreiter, S.: Gans
  trained by a two time-scale update rule converge to a local nash equilibrium.
  Advances in neural information processing systems  \textbf{30} (2017)

\bibitem{lstm}
Hochreiter, S., Schmidhuber, J.: Long short-term memory. Neural computation
  \textbf{9}(8),  1735--1780 (1997)

\bibitem{juefei2022countering}
Juefei-Xu, F., Wang, R., Huang, Y., Guo, Q., Ma, L., Liu, Y.: Countering
  malicious deepfakes: Survey, battleground, and horizon. International Journal
  of Computer Vision pp. 1--57 (2022)

\bibitem{tlvqm}
Korhonen, J.: Two-level approach for no-reference consumer video quality
  assessment. IEEE Transactions on Image Processing  \textbf{28}(12),
  5923--5938 (2019)

\bibitem{human-vs-machines}
Korshunov, P., Marcel, S.: Deepfake detection: humans vs. machines. arXiv
  preprint arXiv:2009.03155  (2020)

\bibitem{faceshifter}
Li, L., Bao, J., Yang, H., Chen, D., Wen, F.: Advancing high fidelity identity
  swapping for forgery detection. In: Proceedings of the IEEE/CVF Conference on
  Computer Vision and Pattern Recognition (CVPR). pp. 5074--5083 (2020)

\bibitem{mos-net}
Lo, C.C., Fu, S.W., Huang, W.C., Wang, X., Yamagishi, J., Tsao, Y., Wang, H.M.:
  {MOSNet: Deep Learning-Based Objective Assessment for Voice Conversion}. In:
  Proc. Interspeech 2019. pp. 1541--1545 (2019).
  \doi{10.21437/Interspeech.2019-2003}

\bibitem{brisque}
Mittal, A., Moorthy, A.K., Bovik, A.C.: No-reference image quality assessment
  in the spatial domain. IEEE Transactions on Image Processing
  \textbf{21}(12),  4695--4708 (2012)

\bibitem{synface}
Nightingale, S., Agarwal, S., H{\"a}rk{\"o}nen, E., Lehtinen, J., Farid, H.:
  Synthetic faces: how perceptually convincing are they? Journal of Vision
  \textbf{21}(9),  2015--2015 (2021)

\bibitem{ai}
Nightingale, S.J., Farid, H.: Ai-synthesized faces are indistinguishable from
  real faces and more trustworthy. Proceedings of the National Academy of
  Sciences  \textbf{119}(8),  e2120481119 (2022)

\bibitem{vggface}
Parkhi, O., Vedaldi, A., Zisserman, A.: Deep face recognition. In: Proceedings
  of the British Machine Vision Conference. pp. 1--12 (2015)

\bibitem{dfgc}
Peng, B., Xiang, W., Jiang, Y., Wang, W., Dong, J., Sun, Z., Lei, Z., Lyu, S.:
  Dfgc 2022: The second deepfake game competition. In: 2022 IEEE International
  Joint Conference on Biometrics (IJCB). pp. 1--10 (2022)

\bibitem{deepfacelab}
Perov, I., Gao, D., Chervoniy, N., Liu, K., Marangonda, S., Um{\'e}, C., Dpfks,
  M., Facenheim, C.S., RP, L., Jiang, J., et~al.: Deepfacelab: Integrated,
  flexible and extensible face-swapping framework. arXiv preprint
  arXiv:2005.05535  (2020)

\bibitem{ff++}
Rossler, A., Cozzolino, D., Verdoliva, L., Riess, C., Thies, J., Nie{\ss}ner,
  M.: Faceforensics++: Learning to detect manipulated facial images. In:
  Proceedings of the IEEE/CVF International Conference on Computer Vision
  (ICCV). pp. 1--11 (2019)

\bibitem{vbliinds}
Saad, M.A., Bovik, A.C., Charrier, C.: Blind prediction of natural video
  quality. IEEE Transactions on Image Processing  \textbf{23}(3),  1352--1365
  (2014)

\bibitem{seshadrinathan2010study}
Seshadrinathan, K., Soundararajan, R., Bovik, A.C., Cormack, L.K.: Study of
  subjective and objective quality assessment of video. IEEE transactions on
  Image Processing  \textbf{19}(6),  1427--1441 (2010)

\bibitem{vgg}
Simonyan, K., Zisserman, A.: Very deep convolutional networks for large-scale
  image recognition. In: Proceedings of the 3rd International Conference on
  Learning Representations (2015), \url{http://arxiv.org/abs/1409.1556}

\bibitem{tian2022generalized}
Tian, Y., Ni, Z., Chen, B., Wang, S., Wang, H., Kwong, S.: Generalized visual
  quality assessment of gan-generated face images. arXiv preprint
  arXiv:2201.11975  (2022)

\bibitem{videval}
Tu, Z., Wang, Y., Birkbeck, N., Adsumilli, B., Bovik, A.C.: Ugc-vqa:
  Benchmarking blind video quality assessment for user generated content. IEEE
  Transactions on Image Processing  \textbf{30},  4449--4464 (2021)

\bibitem{RAPIQUE}
Tu, Z., Yu, X., Wang, Y., Birkbeck, N., Adsumilli, B., Bovik, A.C.: Rapique:
  Rapid and accurate video quality prediction of user generated content. IEEE
  Open Journal of Signal Processing  \textbf{2},  425--440 (2021)

\bibitem{ssim}
Wang, Z., Bovik, A.C., Sheikh, H.R., Simoncelli, E.P.: Image quality
  assessment: from error visibility to structural similarity. IEEE transactions
  on image processing  \textbf{13}(4),  600--612 (2004)

\bibitem{fastvqa}
Wu, H., Chen, C., Hou, J., Liao, L., Wang, A., Sun, W., Yan, Q., Lin, W.:
  Fast-vqa: Efficient end-to-end video quality assessment with fragment
  sampling. Proceedings of European Conference of Computer Vision (ECCV)
  (2022)

\bibitem{gmlog}
Xue, W., Mou, X., Zhang, L., Bovik, A.C., Feng, X.: Blind image quality
  assessment using joint statistics of gradient magnitude and laplacian
  features. IEEE Transactions on Image Processing  \textbf{23}(11),  4850--4862
  (2014)

\bibitem{mosa-net}
Zezario, R.E., Fu, S.W., Chen, F., Fuh, C.S., Wang, H.M., Tsao, Y.: Deep
  learning-based non-intrusive multi-objective speech assessment model with
  cross-domain features. IEEE/ACM Transactions on Audio, Speech, and Language
  Processing  \textbf{31},  54--70 (2023). \doi{10.1109/TASLP.2022.3205757}

\bibitem{lpips}
Zhang, R., Isola, P., Efros, A.A., Shechtman, E., Wang, O.: The unreasonable
  effectiveness of deep features as a perceptual metric. In: Proceedings of the
  IEEE Conference on Computer Vision and Pattern Recognition (CVPR) (2018)

\end{thebibliography}
\end{document}